\documentclass[conference]{IEEEtran}
\usepackage{times}

\usepackage[numbers]{natbib}
\usepackage{multicol}
\usepackage[bookmarks=true]{hyperref}
\usepackage{amsmath,amssymb}
\usepackage{gensymb}
\usepackage{graphicx}
\usepackage[export]{adjustbox}
\usepackage{stfloats}
\usepackage{textcomp}
\usepackage[symbol]{footmisc}

\begin{document}

\title{Learning-Free Grasping of Unknown Objects \\ Using Hidden Superquadrics}


\author{ Yuwei Wu$^{1}$ \quad Weixiao Liu$^{1, 2}$ \quad Zhiyang Liu$^{1}$ \quad Gregory S. Chirikjian$^{1*}$\\
$^1$National University of Singapore \quad $^2$Johns Hopkins University\\
{\tt\small \{yw.wu, mpewxl, mpezyl, mpegre\}@nus.edu.sg}
}


%

\maketitle
\footnotetext[1]{Corresponding author}

\begin{abstract}
Robotic grasping is an essential and fundamental task and has been studied extensively over the past several decades.
Traditional work analyzes physical models of the objects and computes force-closure grasps.
Such methods require pre-knowledge of the complete 3D model of an object, which can be hard to obtain.
Recently with significant progress in machine learning, data-driven methods have dominated the area.
Although impressive improvements have been achieved, those methods require a vast amount of training data and suffer from limited generalizability.
In this paper, we propose a novel two-stage approach to predicting and synthesizing grasping poses directly from the point cloud of an object without database knowledge or learning.
Firstly, multiple superquadrics are recovered at different positions within the object, representing the local geometric features of the object surface.
Subsequently, our algorithm exploits the tri-symmetry feature of superquadrics and synthesizes a list of antipodal grasps from each recovered superquadric.
An evaluation model is designed to assess and quantify the quality of each grasp candidate.
The grasp candidate with the highest score is then selected as the final grasping pose.
We conduct experiments on isolated and packed scenes to corroborate the effectiveness of our method.
The results indicate that our method demonstrates competitive performance compared with the state-of-the-art without the need for either a full model or prior training.

\end{abstract}

\IEEEpeerreviewmaketitle

\section{Introduction}
For the past decades, we have witnessed rapid development and massive success in the field of robotics. Industrial robots are pretty mature and already heavily deployed in industry \cite{appleton2012industrial}. 
In addition, domestic robots have achieved significant success in various aspects of human life \cite{bogue2017domestic}. 
One of the most promising applications for domestic robots is caring for elderly or disabled people. 
However, to safely deploy such nursing-care robots in daily life, there still remain several challenges, one of which is robotic grasping. 

Grasping is an essential and fundamental operation involved in various robotic tasks, such as fetching, housekeeping, cooking, etc. 
Researchers have been committed to tackling this problem. 
Traditional work such as \cite{ding2000computing,liu1999qualitative,ponce1993characterizing, zhu2004planning} proposes analytical methods based on the mathematical and physical models of the objects. 
But their approaches require pre-knowledge of the full 3D model of the objects, which can rarely be satisfied in real-life scenarios. 
Recently with great progress in deep learning, researchers have started taking advantage of big data and training deep neural networks to predict grasp poses given visual inputs \cite{mahler2017dex,morrison2018closing, morrison2020learning,ten2017grasp}.
Those methods have greatly improved the performance to an impressive level.
They can clear a densely packed scene by grasping one object after another.
However, deep learning-based methods require a vast amount of data and computing resources and can suffer from generalization issues.
\begin{figure}[t]
    \centering
    \includegraphics[width=0.98\columnwidth]{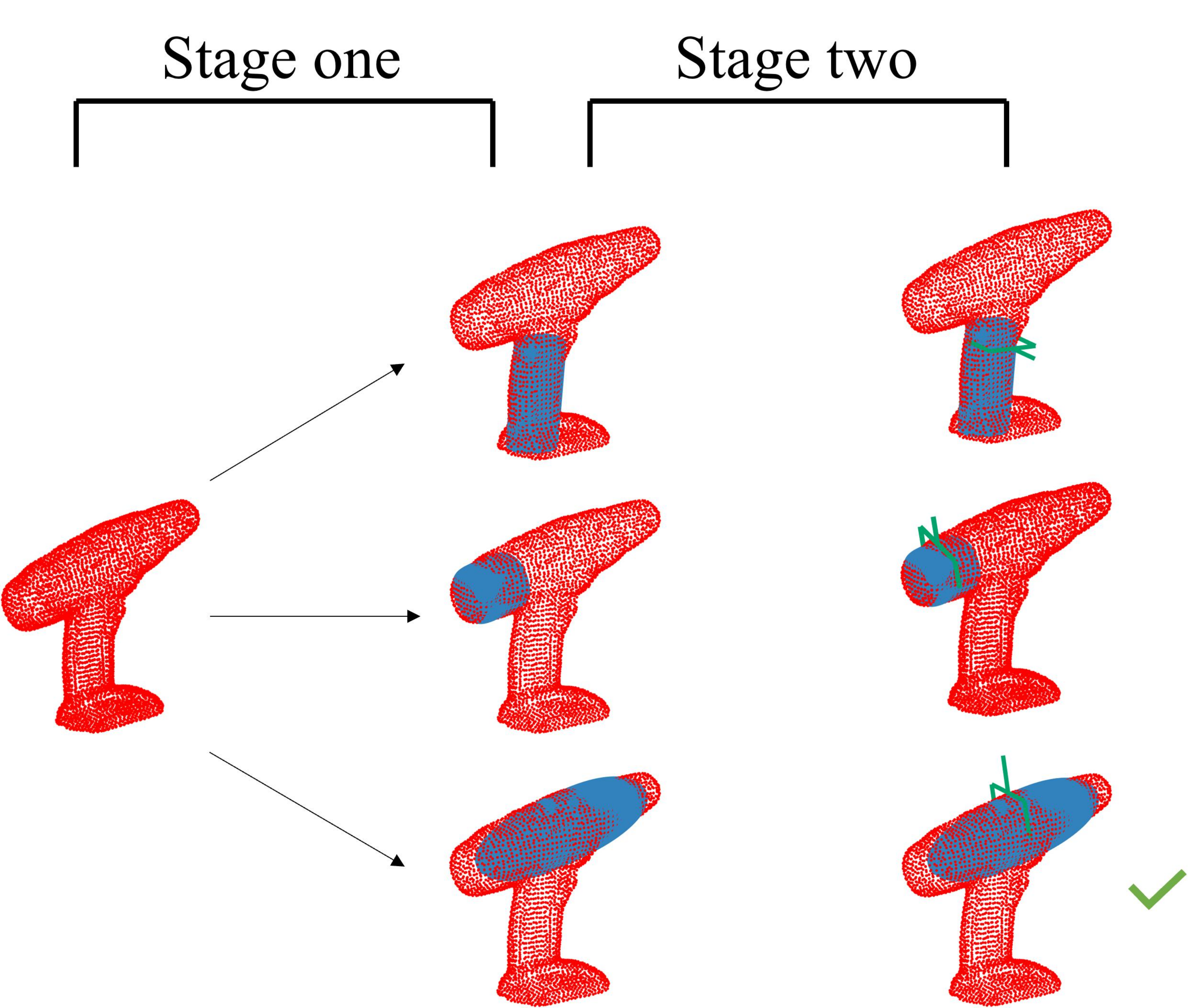}
    \caption{Demonstration of the two-stage approach on a power drill. The algorithm first recovers several hidden superquadrics from the object. Subsequently, multiple grasping poses are synthesized from those superquadrics and the evaluation model selects the most feasible grasp.}
    \label{fig:intro}
\end{figure}
On the other hand, humans can grasp a novel object within seconds of seeing it.
In \cite{biederman1987recognition}, the authors explain that the human visual system extracts geometric primitives and uses them to identify an object.
Furthermore, when humans attempt to grasp an object, how humans choose to grasp depends on part of the shape of the object \cite{cutkosky1990human}.
Therefore to learn from the philosophy of human perception and grasping, we propose a novel two-stage approach for detecting valid and stable grasping poses directly from point clouds of previously unseen objects.
During the first stage, our algorithm identifies and pinpoints several geometric primitives at different positions within the object that are suitable for a parallel-jaw gripper to grasp (Fig \ref{fig:intro}).
We call those geometric primitives hidden primitives.
Superquadrics \cite{barr1981superquadrics} are a family of tri-symmetrical surfaces that include many common shapes like cuboids, cylinders, ellipsoids, and shapes in between.
Not only are they expressive, but the symmetry also guarantees antipodal grasps when applying forces along three principal axes. 
This is beneficial and illuminating to the second stage, where grasping poses are synthesized.
Therefore, we choose the superquadric as our geometric primitive.
Recent works \cite{Liu_2022_CVPR,Liu2023CVPR,paschalidou2019superquadrics,paschalidou2020learning, Wu_2022_eccv} have made a huge success in recovering one or multiple superquadrics from point clouds/meshes/SDFs. 
In \cite{paschalidou2019superquadrics,paschalidou2020learning}, the authors train a neural network to parse 3D objects into superquadric representations without supervision, whereas in \cite{Liu_2022_CVPR,Liu2023CVPR, Wu_2022_eccv}, the authors build a probabilistic model to infer superquadric abstraction directly from point clouds or SDFs without deep learning.
Likewise, our approach builds a similar probabilistic model capable of recovering multiple superquadrics at different positions inside the object. 
In the second stage, based on the recovered hidden primitives, our algorithm synthesizes a series of grasping poses and quantifies the quality of each grasping candidate, and then picks out the most feasible and stable grasping pose from the list of candidates.
We conduct experiments on various isolated objects and packed scenes to validate our approach. 
The experimental results demonstrate that our approach exhibits competitive performance when grasping novel objects. 

Our paper makes several key contributions. 
Firstly, we propose a novel two-stage grasping strategy inspired by how humans perceive and grasp.
Additionally, we adapt the probabilistic models proposed in \cite{Liu_2022_CVPR,Liu2023CVPR, Wu_2022_eccv} to the field of robotic grasping, which can resist noisy points resulting from sensors and calibration errors.
Furthermore, we present a simple, straightforward, yet effective method to synthesize collision-free grasping poses from hidden superquadrics and design an evaluation model to assess the quality of each candidate pose.
Although the method is customized for parallel-jaw grippers, the idea can be extended to different types of grippers and hands. 

\section{Related Work}
This section will briefly discuss the basics of superquadrics, recent work on 3D shape abstraction, and review some relevant work on robotic grasping.
\begin{figure}
    \centering
    \includegraphics[width=0.99\columnwidth]{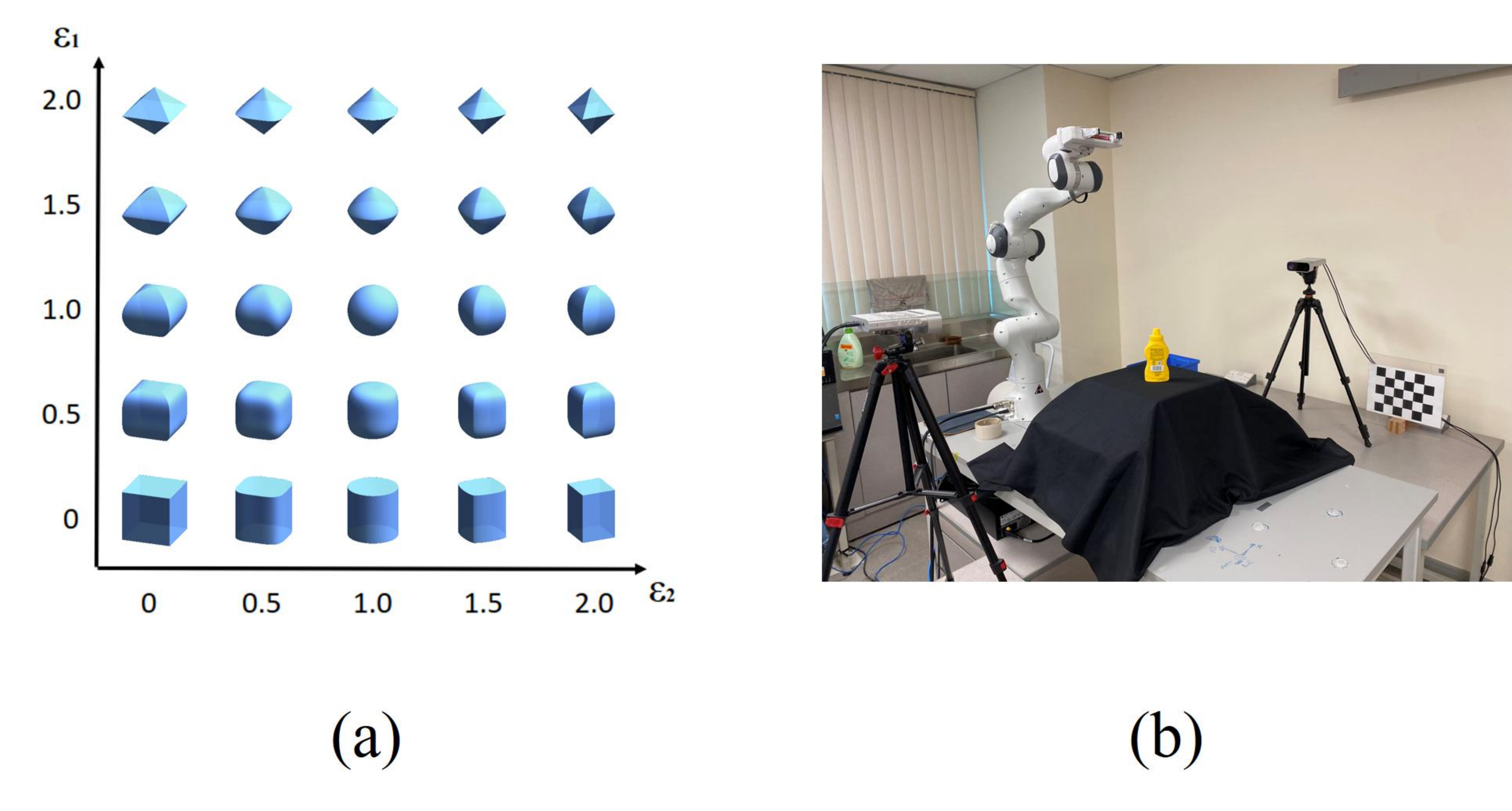}
    \caption{(a) Shape vocabulary of superquadrics \cite{Liu_2022_CVPR}. (b) The setup of the robot arm and sensors.}
    \label{fig:sq}
\end{figure}
\subsection{Superquadrics}
Superquadrics are a family of volumetric surfaces introduced into the computer vision field in \cite{barr1981superquadrics}. 
A superquadric surface can be parameterized by $\omega\in(-\pi,\pi]$ and $\eta\in[-\frac{\pi}{2},\frac{\pi}{2}]$:
\begin{equation}
\begin{aligned}
\label{eq:sq}
&\mathbf{p}(\eta, \omega)=\left[\begin{array}{c}
C_{\eta}^{\varepsilon_{1}} \\
a_{z} S_{\eta}^{\varepsilon_{1}}
\end{array}\right] \otimes\left[\begin{array}{l}
a_{x} C_{\omega}^{\varepsilon_{2}} \\
a_{y} S_{\omega}^{\varepsilon_{2}}
\end{array}\right]=\left[\begin{array}{c}
a_{x} C_{\eta}^{\varepsilon_{1}} C_{\omega}^{\varepsilon_{2}} \\
a_{y} C_{\eta}^{\varepsilon_{1}} S_{\omega}^{\varepsilon_{2}} \\
a_{z} S_{\eta}^{\varepsilon_{1}}
\end{array}\right] \\
&C_{\alpha}^{\varepsilon} \triangleq \operatorname{sgn}(\cos (\alpha))|\cos (\alpha)|^{\varepsilon},\, S_{\alpha}^{\varepsilon} \triangleq \operatorname{sgn}(\sin (\alpha))|\sin (\alpha)|^{\varepsilon},
\end{aligned}
\end{equation}
where $\otimes$ denotes the spherical product \cite{barr1981superquadrics}, $\varepsilon_1$ and $\varepsilon_2$ define the sharpness of the shape, and $a_x$, $a_y$, and $a_z$ control the size and aspect ratio (Fig. \ref{fig:sq}).
The right side of $\otimes$ is a superellipse curve encoded by $a_x,a_y, \text{and} \ \varepsilon_2$, which is the intersection between the superquadric and the $XY$ plane.
We call it the basic shape of the superquadric. 
To fully describe a superquadric in Euclidean space, we need a total number of 11 parameters $\boldsymbol{\theta} = \{\varepsilon_1,\varepsilon_2,a_x,a_y,a_z,g \mid g\in \text{SE(3)}\}$.
One important feature of superquadrics to notice is that they are tri-symmetrical, meaning that each superquadric is symmetrical along each of the three principal axes which are denoted as $\boldsymbol{\lambda_x},\boldsymbol{\lambda_y}, \text{and} \ \boldsymbol{\lambda_z}$.
Formally, the symmetry group of a general superquadric is the Klein 4-group, consisting of the following elements
\begin{equation}
    \{\mathbb{I}, R_x(\pi), R_y(\pi), R_z(\pi) \},
\end{equation}
where $\mathbb{I}$ is the identity element, $R_x(\pi), R_y(\pi), \text{and} \ R_z(\pi)$ are rotations by $\pi$ around $\boldsymbol{\lambda_x},\boldsymbol{\lambda_y}, \text{and} \ \boldsymbol{\lambda_z}$, respectively.
Therefore, each pair of endpoints along each principal axis forms a set of antipodal points. 
And by applying a parallel gripper on the antipodal points, we will obtain antipodal grasps, which guarantees force-closure \cite{chen1993finding}.

\subsection{Shape Abstraction}
Apart from low-level representations of 3D shapes such as voxels \cite{curless1996volumetric}, meshes \cite{lorensen1987marching}, and point clouds \cite{fan2017point}, volumetric representations via geometric primitives have attracted attention, as well.
The superquadric is one of the geometric primitives that has been studied extensively.
In \cite{chevalier2003segmentation, leonardis1997superquadrics}, the authors first introduce a segment-and-fit model that parses an object into different parts and then fits each part with a single superquadric.
Although that method of segmentation is intuitive, it suffers from limited accuracy.
More recently, researchers have exploited the advantage of big data and deep neural networks to predict abstractions from RGB images or meshes \cite{paschalidou2019superquadrics, paschalidou2020learning}.
On the other hand, the authors in \cite{Liu_2022_CVPR, Wu_2022_eccv} propose probabilistic models to infer shape abstractions directly from point clouds.
Optimization with a strategy for avoiding local minima is employed in their algorithms, leading to more accurate results.

\subsection{Robotics Grasping}
Robotic grasping is an essential task in robotics and has been studied extensively. In the early days, researchers analyze the mathematical and physical models of the objects and synthesize force-closure grasping poses \cite{ding2000computing,liu1999qualitative,ponce1993characterizing, zhu2004planning}.
The major drawback lies within the assumption of pre-knowledge of objects' geometric information.
Such an assumption is rarely satisfied in real-world applications.
In contrast, empirical approaches learn from available grasping results and apply the learned knowledge to novel objects. 
In \cite{pelossof2004svm, saxena2008robotic}, the authors build a regression map between input shapes and grasping quality, and detect grasping poses on unseen objects.
With significant progress in deep learning, data-driven methods have recently dominated robotic grasping.
Sampling approaches first sample a series of grasp candidates and then estimate the quality of each individual grasp via a trained neural network \cite{kokic2020learning, liang2019pointnetgpd, mousavian20196, murali20206, ten2017grasp}. 
Direct regression, on the other hand, learns a deep network to predict high-quality grasps directly from visual inputs \cite{breyer2021volumetric,fang2020graspnet,sundermeyer2021contact}.
To reduce the complexity of 6D grasping, some researchers propose simplifying the full 6D grasping to 4D planar grasping, i.e. the gripper can only grasp the object from the top-down direction \cite{kumra2020antipodal,mahler2017dex,morrison2018closing, morrison2020learning,zhu2022sample}. 
Although planar grasps have reached astonishing achievements, 4D methods are less competent compared to full 6D grasps due to the loss of two degrees of freedom.
A more comprehensive review can be found in \cite{newbury2022deep}.
Researchers have also studied predicting grasps from primitives.
In \cite{el2007learning, sahbani2009hybrid}, the authors proposed a grasping-by-components strategy. 
They first segment the object model into several parts, each of which is presented by a tapered and bent superquadric.
Then by supervised training, they learn how to grasp each deformed superquadric.
However, their methods require an exact and complete 3D model of an object and an accurate segmentation algorithm, both of which are impractical.
In \cite{makhal2018grasping, vezzani2017grasping}, the authors proposed a grasping method relying on a single superquadric representation of objects.
But in reality, most objects cannot be fairly represented by a single superquadric, resulting in a decrease in performance due to inaccurate shape abstraction.

\section{Methods}

\begin{figure*}[t]
    \centering
    \includegraphics[width=1.99\columnwidth]{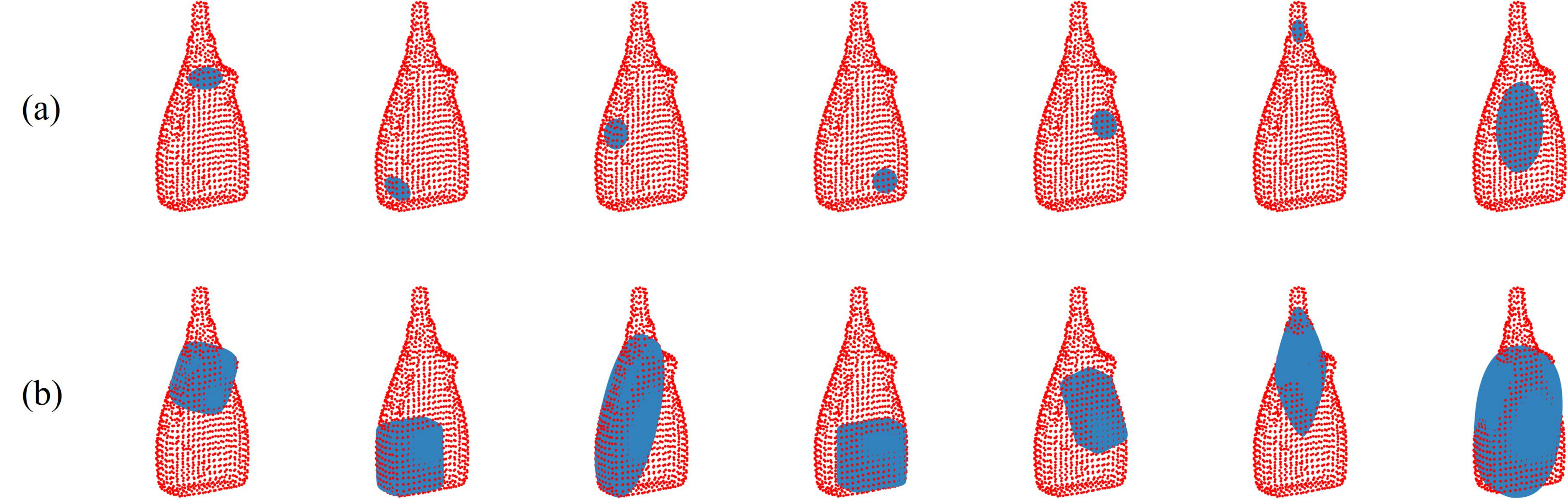}
    \caption{(a) 6+1 superquadrics at different positions are initialized. (b) The corresponding recovered superquadrics after optimization.}
    \label{fig:stage one}
\end{figure*}

We consider the problem of 6-DoF grasp detection from point clouds of unseen objects. We assume a robot arm equipped with a parallel-jaw gripper mounted on a planar tabletop. The point cloud of an object is acquired by two calibrated Azure Kinect depth sensors fixed on the two sides of the table. 
Each depth sensor will capture a partial point cloud of the object and we fuse them together to obtain a relatively complete point cloud.
The complete setup is shown in Fig. \ref{fig:sq}(b).
Given the point cloud of an object, we propose a novel two-stage approach to detecting and synthesizing valid grasping poses.

\subsection{Stage One: Hidden Superquadric Recovery}
At this stage, the goal is to find the hidden superquadric inside the object. In other words, the algorithm tries to identify which part of a given object can be approximately represented by a single superquadric and then obtain the parameter of that superquadric. 
There are two key observations: 1) in most cases a single superquadric cannot fairly represent the full object; 2) when making a grasp, humans only need to identify the graspable part of an object and ignore the others.
Formally, given a point cloud $\boldsymbol{X}$, we need to find a subset $\boldsymbol{Y} \subset \boldsymbol{X}$ which can be fairly represented by a superquadric surface $\boldsymbol{\mathcal{S}}_{\boldsymbol{\theta}}$. 

Following the work in \cite{Liu_2022_CVPR}, our algorithm assumes a workspace $\mathbb{V}$ encapsulating the whole point set with volume $V$.
Subsequently, each element $\mathbf{x} \in \boldsymbol{X}$ is generated either from a outlier component $p_o(\mathbf{x})$ or from a superquadric component $\mathcal{N}(\mathbf{x}|\boldsymbol{\mu}, \mathbf{\Sigma})$, where
\begin{equation}
    p_o(\mathbf{x})= \begin{cases}
        \frac{1}{V} & \text{if } \mathbf{x} \in \mathbb{V} \\
        0 & \text{if } \mathbf{x} \notin \mathbb{V},
        \end{cases}
\end{equation}
and $\mathcal{N}(\mathbf{x}|\boldsymbol{\mu}, \mathbf{\Sigma})$ is a Gaussian distribution with mean $\boldsymbol{\mu}$ being a random point on a superquadric surface $\boldsymbol{\mathcal{S}}_{\boldsymbol{\theta}}$ and covariance $\mathbf{\Sigma}$ representing the noise level. Introducing a latent variable $z$ indicating the membership of each element gives:
\begin{equation}
    \begin{gathered}
     p(\mathbf{x}|\boldsymbol{\mu}, \mathbf{\Sigma}, z) = p_o(\mathbf{x})^{1-z}\cdot\mathcal{N}(\mathbf{x}|\boldsymbol{\mu}, \mathbf{\Sigma})^{z}\\
     z\sim p(z)=\mathrm {Bernoulli}(1-w_o),
    \end{gathered}
\label{eqn_gum_latent}
\end{equation}
where $w_o$ denotes the probability of a point sampled from the outlier component.
In other words, we cast the superquadric recovery as a mixture model problem. 
All points within the workspace are either sampled from the outlier component or the hidden superquadric.
To infer the posterior distributions of latent variable $z$ and the superquadric parameter $\boldsymbol{\theta}$, the optimization-based Gibbs sampling strategy proposed in \cite{Wu_2022_eccv} is applied.
Initialization of the superquadric will significantly affect the final recovered result.
And since multiple different parts of an object can be represented by a superquadric, multiple superquadrics at different positions are initialized.
Specifically, the point set $\boldsymbol{X}$ is parsed into $K$ parts $\{\boldsymbol{\Phi}_1,\boldsymbol{\Phi}_2,...,\boldsymbol{\Phi}_K\}$ via the K-means algorithm. 
Then, for each subset $\boldsymbol{\Phi}_i$, an ellipsoid $\boldsymbol{\theta}_i$ is initialized whose moment-of-inertial (MoI) is two times smaller than the MoI of $\boldsymbol{\Phi}_i$.
To account for the situation in which $\boldsymbol{X}$ itself can be represented by a single superquadric, an extra ellipsoid $\boldsymbol{\theta}_{extra}$ is initialized whose moment-of-inertial (MoI) is two times smaller than the MoI of $\boldsymbol{X}$.
We choose the initial superquadric to be ellipsoid as in \cite{Liu_2022_CVPR,solina1990recovery,Wu_2022_eccv}.
Thus, a total of $K+1$ ellipsoids are initialized from which the sampling process starts one by one, and a total number of $K+1$ optimized superquadrics will be acquired.
An illustration is shown in Fig \ref{fig:stage one}. 
As we can observe, different initializations will lead to different hidden superquadrics.
The main takeaway is that we consider the majority of $\boldsymbol{X}$ to be outliers by assuming a high value of $w_o$ and then each initialized ellipsoid will try to evolve and fit into the object locally.

\subsection{Stage Two: Grasp Synthesis and Evaluation}
\subsubsection{Gripper}
\begin{figure}[h]
    \centering
    \includegraphics[width=0.9\columnwidth]{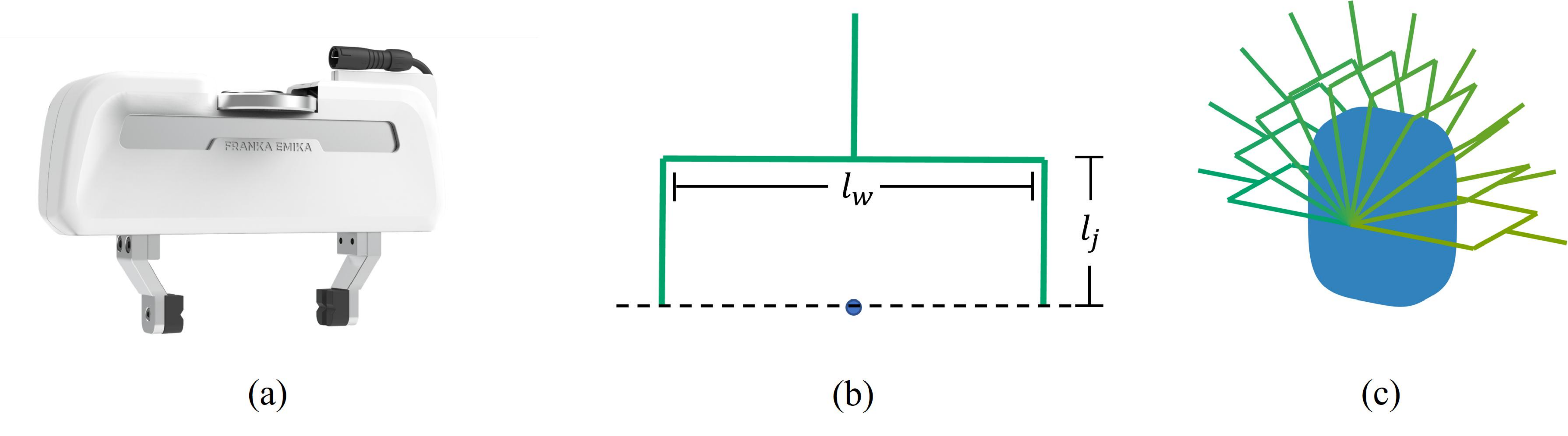}
    \caption{(a) The physical model of the Franka Emika\texttrademark \ hand. (b) The simplified model of Franka hand. We denote the length of the jaw as $l_j$, the maximum width of the gripper as $l_w$, the closing line of the gripper as $\boldsymbol{\lambda}$, and the closing point of the gripper as $\boldsymbol{P_G}$. (c) Infinitely many grasping poses can be generated along one principal axis of the superquadric by rotating the gripper.}
    \label{fig:gripper}
\end{figure}

Throughout this work, we use a parallel-jaw gripper. 
The physical model of a gripper is simplified as shown in Fig. \ref{fig:gripper}.
We denote the length of the jaw as $l_j$ and the maximum width of the gripper as $l_w$. 
The line connecting the two endpoints of the jaws is called the closing line of the gripper, denoted as $\boldsymbol{\lambda}$ (marked as the black dotted line in the figure). The middle point of the two endpoints of the two jaws is called the closing point of the gripper, denoted as $\boldsymbol{P_G}$ (marked as the blue point in the figure).

\subsubsection{Grasp Synthesis}
An empty set $\mathcal{G} = \emptyset$ is initialized as the grasping pose candidate set.
Due to the tri-symmetry feature of superquadrics, each pair of endpoints along each principal axis forms a set of antipodal points.
Ideally, given a superquadric $\boldsymbol{\mathcal{S}}_{\boldsymbol{\theta}}$ recovered in stage one, a list of antipodal grasps can be obtained by moving the parallel gripper to the poses where the closing line $\boldsymbol{\lambda}$ of the gripper overlaps with one of the principal axes of the superquadric.
And by rotating the gripper with respect to $\boldsymbol{\lambda}$, infinitely many grasping poses can be obtained (Fig. \ref{fig:gripper}(c)).
In our work, a grasping pose at every $10^\circ$ is sampled. 
All poses sampled by rotating the gripper around $\boldsymbol{\lambda}$ form a set $\Delta_{\boldsymbol{\lambda}}$, which is added to $\mathcal{G}= \mathcal{G} \cup \Delta_{\boldsymbol{\lambda}}$.
Normally, only three sets $\Delta_{\boldsymbol{\lambda_x}}, \Delta_{\boldsymbol{\lambda_y}}, \text{and} \ \Delta_{\boldsymbol{\lambda_z}}$ of antipodal grasps can be synthesized given a superquadric.
However, depending on the shape of the hidden superquadric, there could be more antipodal grasps.
\begin{figure}[t]
    \centering
    \includegraphics[width=0.9\columnwidth]{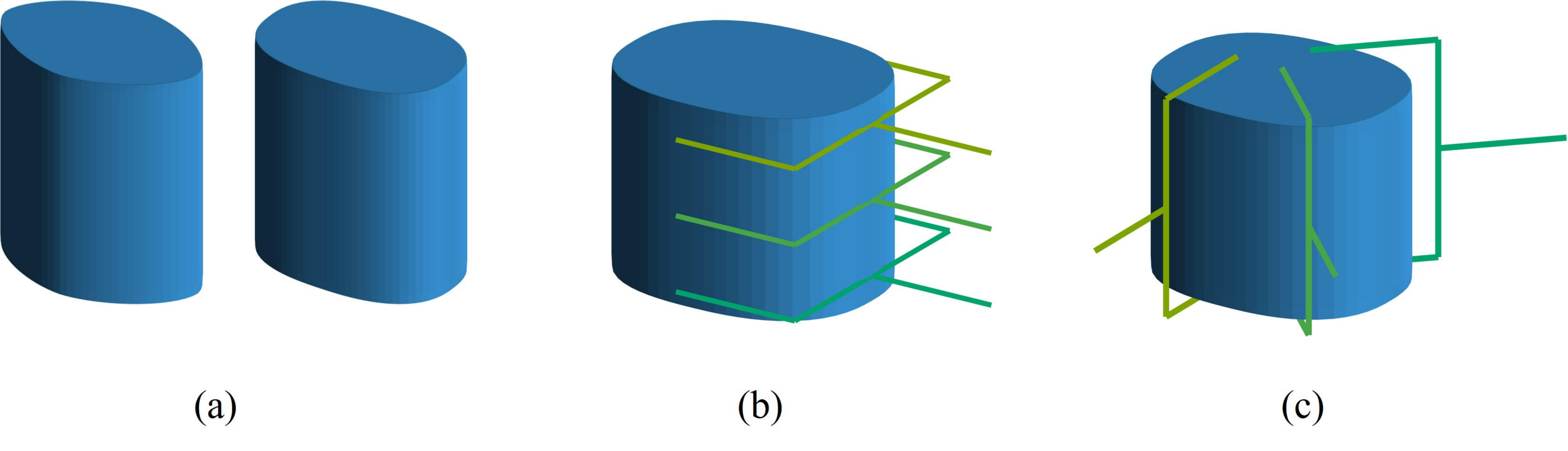}
    \caption{(a) Superquadric Examples when $\varepsilon_1$ approaches 0. (b) Examples of more grasping poses by moving the gripper up and down along the z-axis. (c) Examples of more grasping poses by moving the gripper along the x-axis and y-axis.}
    \label{fig:prism}
\end{figure}

$\boldsymbol{1) \ \varepsilon_1 \rightarrow 0:}$ When $\varepsilon_1$ approaches 0, the shape of a superquadric will become a prism with a superellipse encoded by $\varepsilon_2, a_x, \text{and} \ a_y$ as its base and with height being $2a_z$ (Fig. \ref{fig:prism}(a)).
In this case, apart from $\boldsymbol{\lambda}$ overlapping with $\boldsymbol{\lambda_x}$, the gripper can move up and down along the z-axis of the hidden superquadric and still guarantees antipodal grasps as shown in Fig. \ref{fig:prism}(b).
Therefore, the gripper will move up and down at every 15 millimeters (mm), obtaining a new closing line $\boldsymbol{\lambda_x}^{\prime}$. After that, a new set $\Delta_{\boldsymbol{\lambda_x}^{\prime}}$ of grasping poses is added to the $\mathcal{G}$. 
The same goes for the $\Delta_{\boldsymbol{\lambda_y}^{\prime}}$.
Furthermore, since the two bases are parallel and identical, the gripper can move along the x-axis and y-axis within the base as shown in Fig. \ref{fig:prism}(c).
A 2D grid is generated on the base whose size is 15mm and the gripper will move to every junction point of the grid within the base, obtaining a new closing line $\boldsymbol{\lambda_z}^{\prime}$.
And then, a new set $\Delta_{\boldsymbol{\lambda_z}^{\prime}}$ of grasping poses is added to the $\mathcal{G}$.

\begin{figure}[b]
    \centering
    \includegraphics[width=0.9\columnwidth]{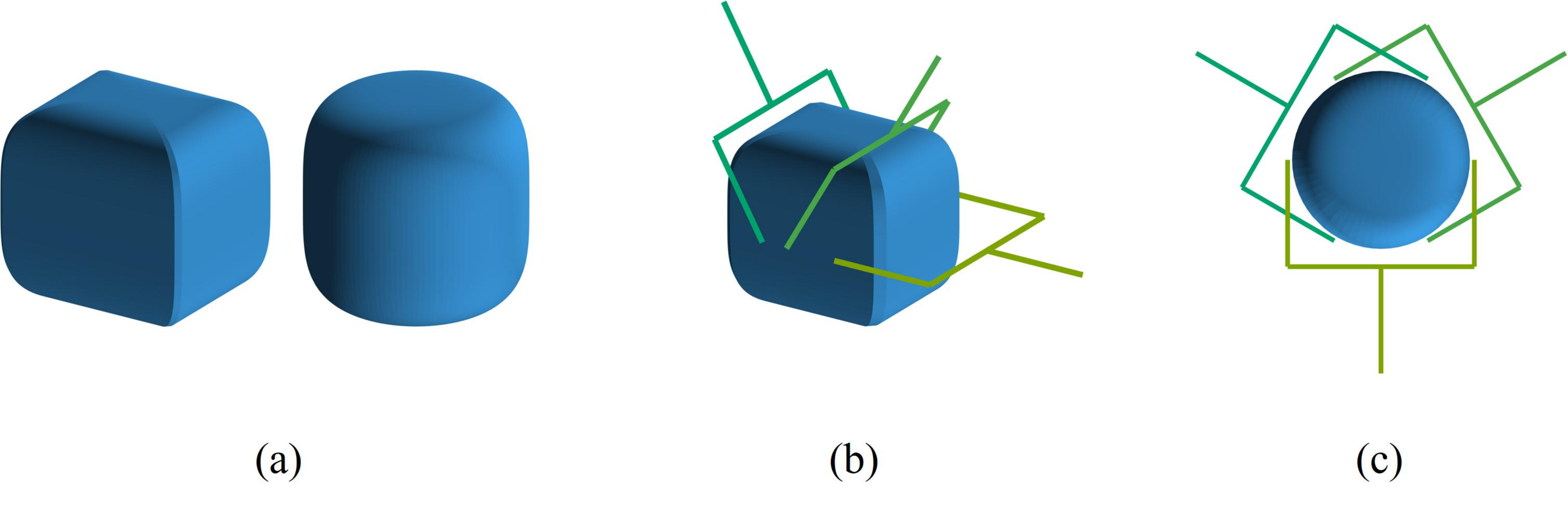}
    \caption{(a) Superquadric Examples with basic shapes being rectangle and circle, respectively. (b) Examples of more grasping poses by moving the gripper along the y-axis when the basic shape of superquadric is a rectangle. (c) Examples of more grasping poses by rotating the gripper around the z-axis when the basic shape of superquadric is a circle.}
    \label{fig:prism2}
\end{figure}
$\boldsymbol{2) \ \varepsilon_2 \rightarrow 0:}$ Recall that the basic shape of the superquadric is the superellipse obtained by the intersection between the superquadric and the $XY$ plane.
In this case, the basic shape of the superquadric is a rectangle.
After aligning $\boldsymbol{\lambda}$ with $\boldsymbol{\lambda_x}$ and obtaining $\Delta_{\boldsymbol{\lambda_x}}$, the gripper can move along the y-axis back and forth to synthesize more grasping poses (Fig. \ref{fig:prism2}(b)).
Again, an interval of 15mm is chosen and the gripper will move to every interval step and obtain additional $\Delta_{\boldsymbol{\lambda_x}^{\prime}}$.
The same goes for the y-axis.

$\boldsymbol{3) \ \varepsilon_2 = 1 \ \text{and} \ a_x=a_y:}$ In this case, the basic shape of the superquadric is a circle. Therefore, the gripper can rotate around the z-axis and still produces antipodal grasps (Fig.\ref{fig:prism2}(c)).
An interval of $\frac{\pi}{8}$ is chosen and the gripper will rotate around the z-axis at each interval, after which $\Delta_{\boldsymbol{\lambda_R}}$ is synthesized and added to the $\mathcal{G}$.

 If the hidden superquadric happens to be the cuboid ($\varepsilon_1 \rightarrow 0 \ \text{and} \ \varepsilon_2 \rightarrow 0$) or cylinder ($\varepsilon_1 \rightarrow 0, \varepsilon_2=1, \text{and} \ a_x=a_y$), then both cases will be taken into consideration.

\begin{figure}[h]
    \centering
    \includegraphics[width=0.7\columnwidth]{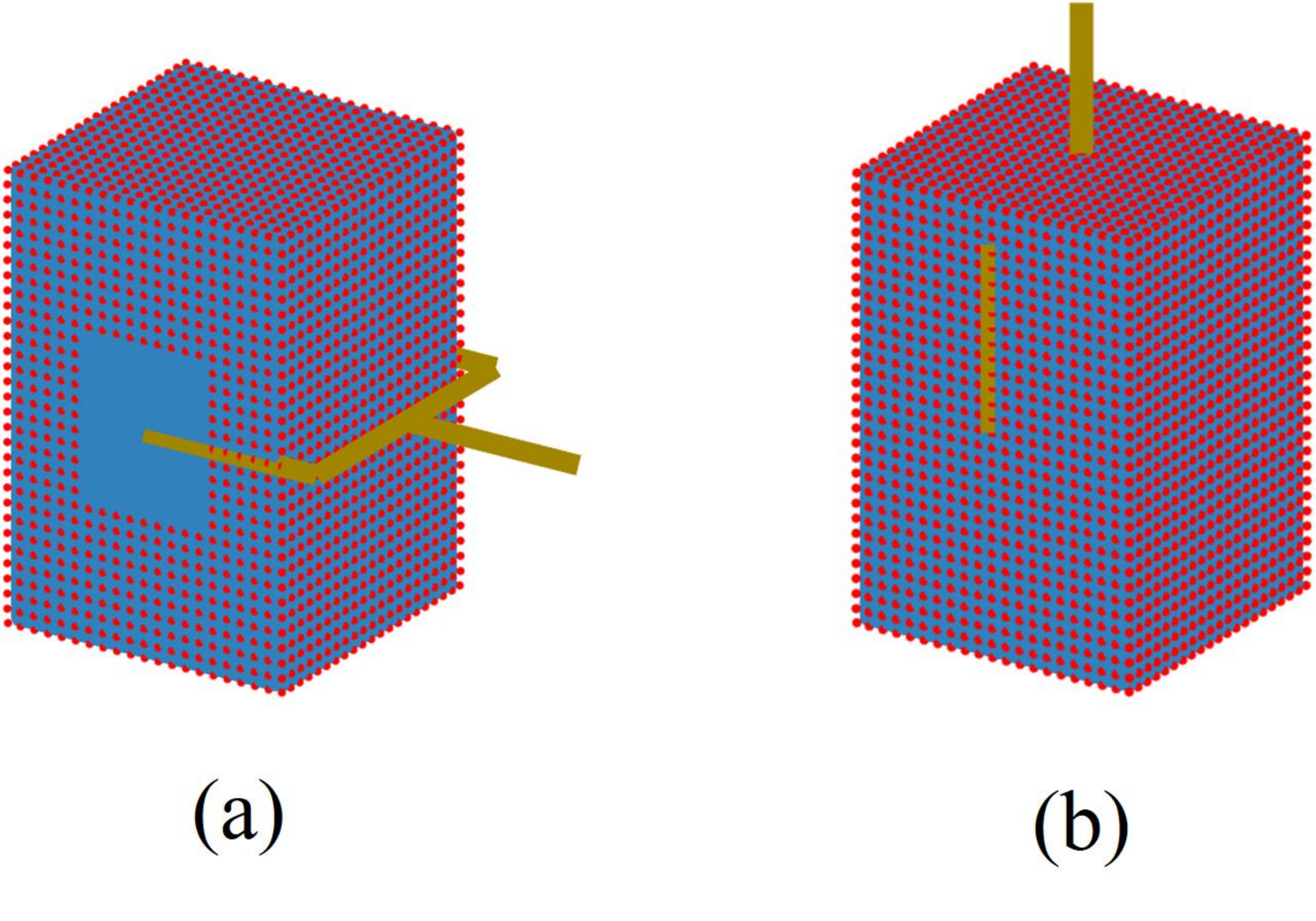}
    \caption{(a) Invalid grasp due to lack of supporting structure. (b) Invalid grasp due to physical collision.}
    \label{fig:invalid}
\end{figure}

However, some poses might be invalid due to the lack of supporting structures or physical collision, which need to be eliminated.
As Fig. \ref{fig:invalid}(a) shows, although the superquadric is a good fit for the object, the left side of the object is missing a part that makes it lack of supporting structure for the gripper to apply force on, which means it is an invalid grasp.
Additionally, the physical collision should also be taken into consideration.
Fig. \ref{fig:invalid}(b) demonstrates one case that the gripper could collide with the physical object due to the length limit of the parallel-jaw.

\begin{figure}[t]
    \centering
    \includegraphics[width=0.8\columnwidth]{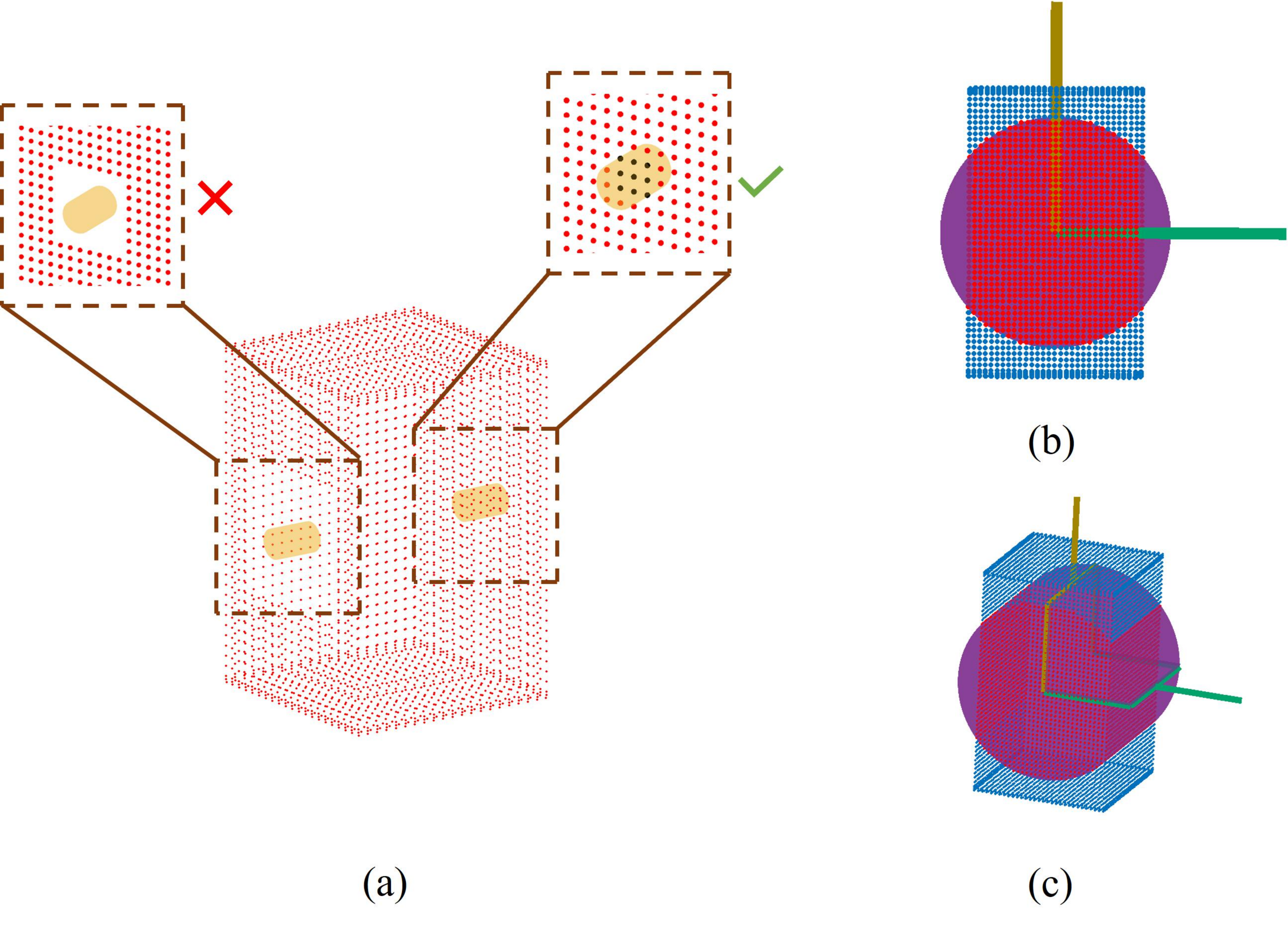}
    \caption{(a) Illustration of supporting structures. (b-c) Illustration of physical collision.}
    \label{fig:eliminate_invalid}
\end{figure}

To determine if there are supporting structures given a grasping pose (without loss of generality, we assume the given pose is from $\Delta_{\boldsymbol{\lambda_x}}$), a small cylinder at both endpoints of the x-axis is generated and our algorithm checks how many points from $\boldsymbol{X}$ are within each cylinder, respectively. 
Unless both numbers exceed a threshold $\boldsymbol{\kappa}$, this pose is excluded from $\mathcal{G}$.
As shown in Fig. \ref{fig:eliminate_invalid}(a), the cylinder generated at the left endpoint contains no points from $\boldsymbol{X}$, which means there is no supporting structure at this position on which the gripper can apply force.
Thus, all poses from $\Delta_{\boldsymbol{\lambda_x}}$ need to be eliminated since the left side has no supporting structure.
Furthermore, to check whether a given pose collides (again we assume the given pose is from $\Delta_{\boldsymbol{\lambda_x}}$), we need to find out all potential colliding points.
Specifically, a cylinder is generated whose radius is $l_j$, height  is $min(a_x, \frac{l_w}{2})$, centroid locates at the centroid of the superquadric, and direction of the height overlaps with the x-axis of $\boldsymbol{\mathcal{S}}_{\boldsymbol{\theta}}$.
Any points that lie within the two base planes of the cylinder and are not at the interior of the cylinder cloud have collided with the gripper, as demonstrated in Fig. \ref{fig:eliminate_invalid}(b-c).
All the red points will not collide with the gripper whereas all the blue points could potentially collide with the gripper due to the length limit of the parallel-jaw and the gripper width.
The purple cylinder is the one used to determine collision points.
The size of the cylinder reflects the size of the gripper.
The green grasp causes no collision while the yellow grasp intersects with the object which is not feasible in the real world.
A pose will be kept in $\mathcal{G}$ only if the posed gripper does not intersect with any of the collided points.
The above elimination process is iterated for every candidate and we will obtain a purged list of $\mathcal{G}$.

\subsubsection{Candidates Evaluation}
In this step, we design an evaluation model to quantify the quality of each candidate $\boldsymbol{{\mathrm{g}}} \in \mathcal{G}$.
The score of each $\boldsymbol{{\mathrm{g}}}$ is calculated as follows:
\begin{equation}
    h = h_\alpha \times h_\beta \times h_\gamma \times h_\delta,
\end{equation}
where the range of each component is $(0,1]$, and so the highest score will be 1 and the lowest score will be 0.
The four components represent the score of the goodness of superquadric, coverage of superquadric, curvature, and distance to the center of mass, respectively.

\textbf{Goodness of Superquadric:} Notice that each synthesized $\boldsymbol{{\mathrm{g}}}$ relies heavily on the accuracy of the corresponding recovered superquadric. 
The more accurately a superquadric fits the geometry of the object, the more confidence we have in those grasping poses generated from it.
To measure the goodness of recovered superquadric, we use the point-to-surface metric \cite{Liu_2022_CVPR} defined as follows:
\begin{equation}
    \alpha(\boldsymbol{Y}, \mathbf{S}) \doteq \frac{1}{N} \sum_{i=1}^{N} \min_{\boldsymbol{s}_j \in \mathbf{S}} \|\boldsymbol{y}_i-\mathbf{s}_j\|_2
    \label{eq:error_metric}
\end{equation}
where $\mathbf{S}=\{\mathbf{s}_1,\mathbf{s}_2,...,\mathbf{s}_M\}$ is a set of points densely and evenly sampled on the recovered superquadric surface $\boldsymbol{\mathcal{S}}_{\boldsymbol{\theta}}$ and $\boldsymbol{Y}=\{\boldsymbol{y}_1,\boldsymbol{y}_2,...,\boldsymbol{y}_N\} \subset \boldsymbol{X}$ is the set of points our algorithm identifies as corresponded to $\boldsymbol{\mathcal{S}}_{\boldsymbol{\theta}}$.
Then, the score of goodness is defined as follows:
\begin{equation}
    h_\alpha = \exp{\left(-\frac{\alpha^2}{q_\alpha}\right)}.
\end{equation}
An exponential model is used to quantify the effect of fitting accuracy with $q_\alpha$ being a hyperparameter that controls the rate of drop.
If the recovered superquadric is a perfect fit, then $\alpha$ will be 0, and $h_\alpha$ will be 1.
Otherwise, the score will drop with an increase in fitting error.

\begin{figure}[h]
    \centering
    \includegraphics[width=0.6\columnwidth]{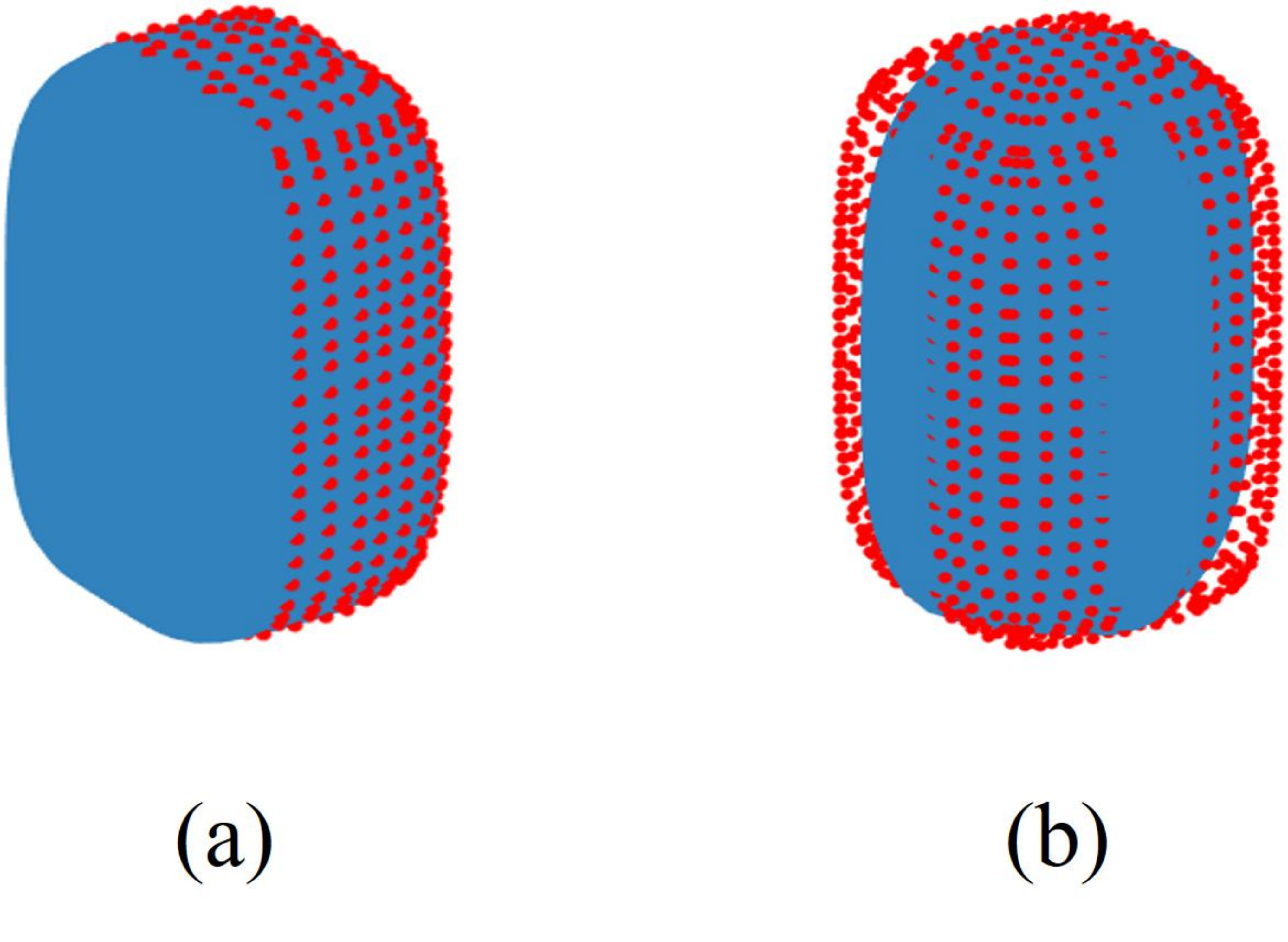}
    \caption{(a) Example of a good fit with bad coverage. (b) Example of good coverage with a bad fit.}
    \label{fig:coverage}
\end{figure}

\textbf{Coverage of Superquadric:} The goodness alone cannot fully determine if a superquadric represents the geometry of the object well.
If the point set $\boldsymbol{Y}$ only covers a small portion of the superquadric, then a well-fitted superquadric may still lead to inaccurate grasps (Fig. \ref{fig:coverage}).
Coverage is computed as follows:
\begin{equation}
    \beta = \frac{|\mathbf{T}|}{\mathbf{|S|}},
\end{equation}
where $\mathbf{T}=\{\mathbf{t}| \mathbf{t} \in \mathbf{S}, \min_{\boldsymbol{y}_j \in \boldsymbol{Y}} \|\boldsymbol{y}_i-\mathbf{t}\|_2\ \leq d_{th}\}$ and $|.|$ denotes the number of the set.
Basically, our algorithm samples evenly and densely from a superquadric and computes how many sampled points are close enough to $\boldsymbol{Y}$, which form a set $\mathbf{T}$.
And we use the ratio between $|\mathbf{T}|$ and $|\mathbf{S}|$ to approximate coverage of superquadric.
The score of coverage is defined as follows:
\begin{equation}
    h_\beta = \beta^2.
\end{equation}

\textbf{Curvature:} For parallel gripper, the more flat the contact part of the object is, the more stable the grasp will be.
Therefore, we compute the average Gaussian curvature $\gamma$ around the endpoints of the superquadric and treat it as a measurement of the flatness of the part of the object that will make contact with the gripper.
Then, the score of curvature is defined as follows:
\begin{equation}
    h_\gamma = \exp{\left(-\frac{\gamma^2}{q_\gamma}\right)}.
\end{equation}
The smaller $\gamma$ is, the flatter the contact surface will be, and the higher score will be earned.

\textbf{Distance to Center of Mass:} To have a more stable grasp, the closing point of the gripper $\boldsymbol{P_G}$ needs to be closer to the center of mass of the object to reduce the torque resulting from gravity.
We assume the density of objects being grasped is uniform and use ${\boldsymbol{\Bar{X}}}$ to approximate the center of mass.
Then the distance $\delta$ between the closing point of the gripper and the center of mass of the object is calculated and
\begin{equation}
    h_\delta = \exp{\left(-\frac{\delta^2}{q_\delta}\right)}.
\end{equation}

Those four components affect the quality of the grasp jointly.
Each candidate $\boldsymbol{\mathrm{g}}$ is evaluated according to the above criteria and our approach will choose the one with the highest score.

\section{Experiments}
\begin{figure}[t]
    \centering
    \includegraphics[width=0.9\columnwidth]{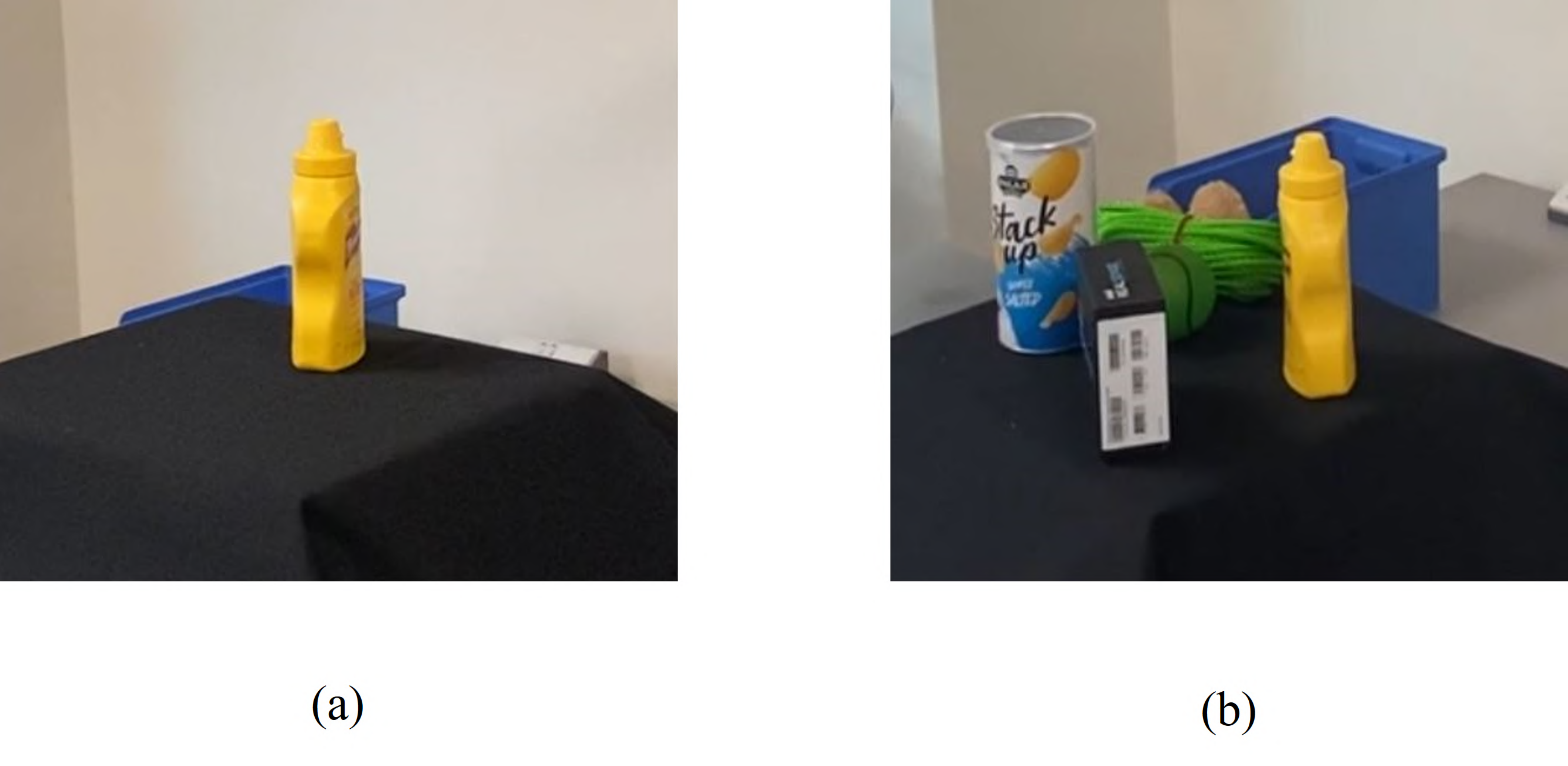}
    \caption{(a) Isolated scene. (b) Packed Scene.}
    \label{fig:scene}
\end{figure}

We conduct experiments on both isolated scenes and packed scenes, where the former means a single object is placed in the workspace for the arm to grasp and the latter means several objects spread out in the workspace such that they are not touching (Fig. \ref{fig:scene}). 
The metric in isolated scenes is \textit{Grasp Success Rate (GSR)} which is defined as the ratio between No. of successful grasps and total No. of attempted grasps.
We define a successful grasp as follows: the robot arm grasps the object and moves it from the workspace to a bin next to the arm without dropping it.
Besides GSR, we also evaluate the \textit{Clearance Rate (CR)} defined as the percentage of objects that are removed from the clutter for the packed scenes.
The number of $K$ in the first stage depends on the number of the point cloud of the object:
\begin{equation}
        K = \begin{cases}
        6 & \text{if } |\boldsymbol{X}| < 8000 \\
        8 + 2 \times \left[\frac{|\boldsymbol{X}|-8000}{4000}\right] & \text{if } |\boldsymbol{X}| \geq 8000.
        \end{cases}
\end{equation}
The more points an object contains, the larger it will be and we assume the more hidden superquadrics are inside the object.
The hyperparameters $q_\alpha, q_\gamma, \text{and} \ q_\delta$ are tuned to be $0.002, 0.5, \text{and} \ 0.005$.
All the algorithms are implemented in Python on a computer equipped with AMD Ryzen\texttrademark \ 9 5950X (3.4GHz).

\subsection{Isolated objects}

\begin{figure}[h]
    \centering
    \includegraphics[width=0.9\columnwidth]{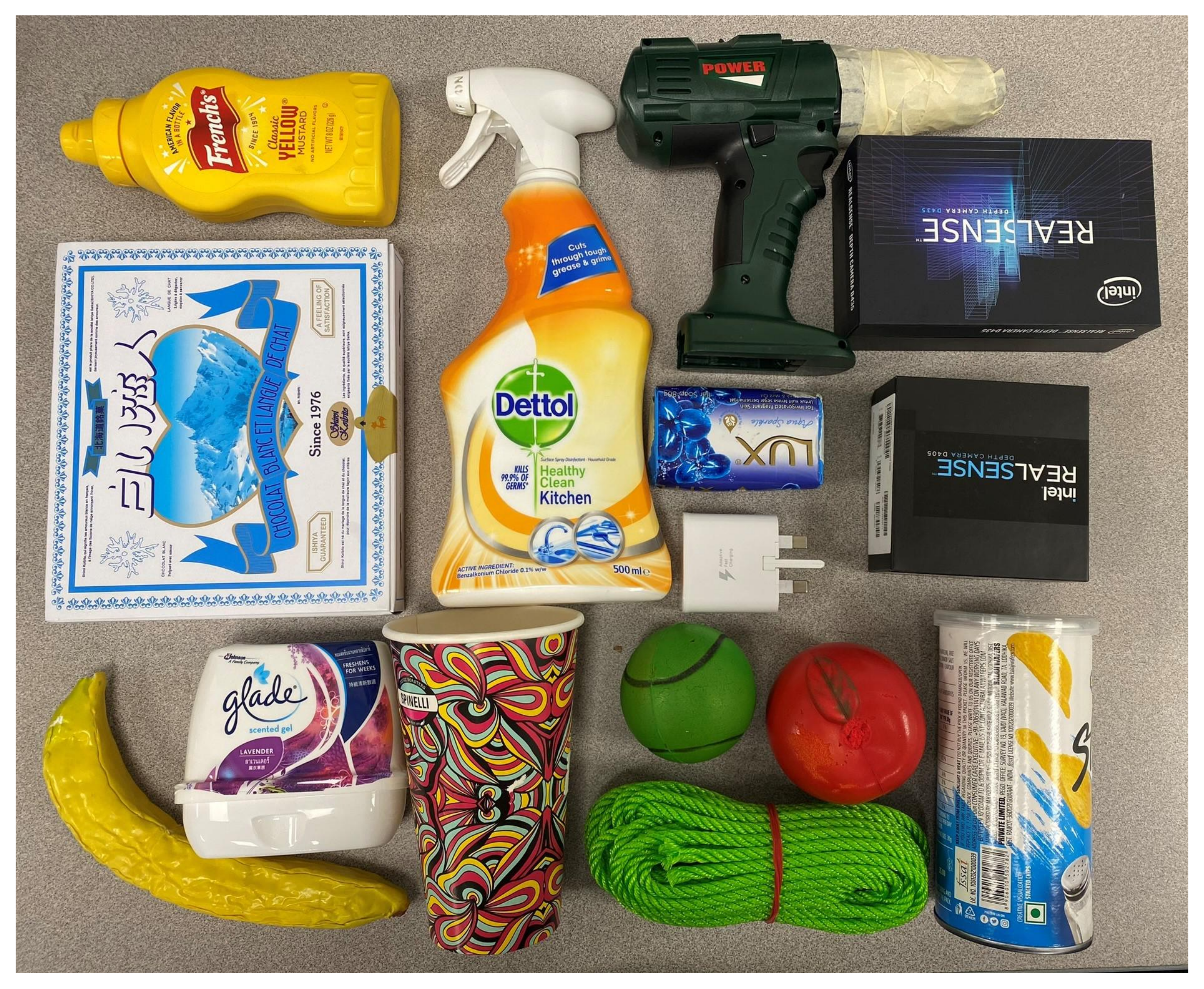}
    \caption{15 objects that are used in our experiment}
    \label{fig:object} 
\end{figure}

In this experiment, we choose 15 objects with various sizes, shapes, and weights as shown in Fig. \ref{fig:object}.
10 attempts are made to grasp each object placed at different poses at each attempt.
Thus, a total number of 150 grasps are attempted. 
During each attempt, the object is placed in a random upright pose within the workspace.

\begin{figure*}[t]
    \centering
    \includegraphics[width=1.9\columnwidth]{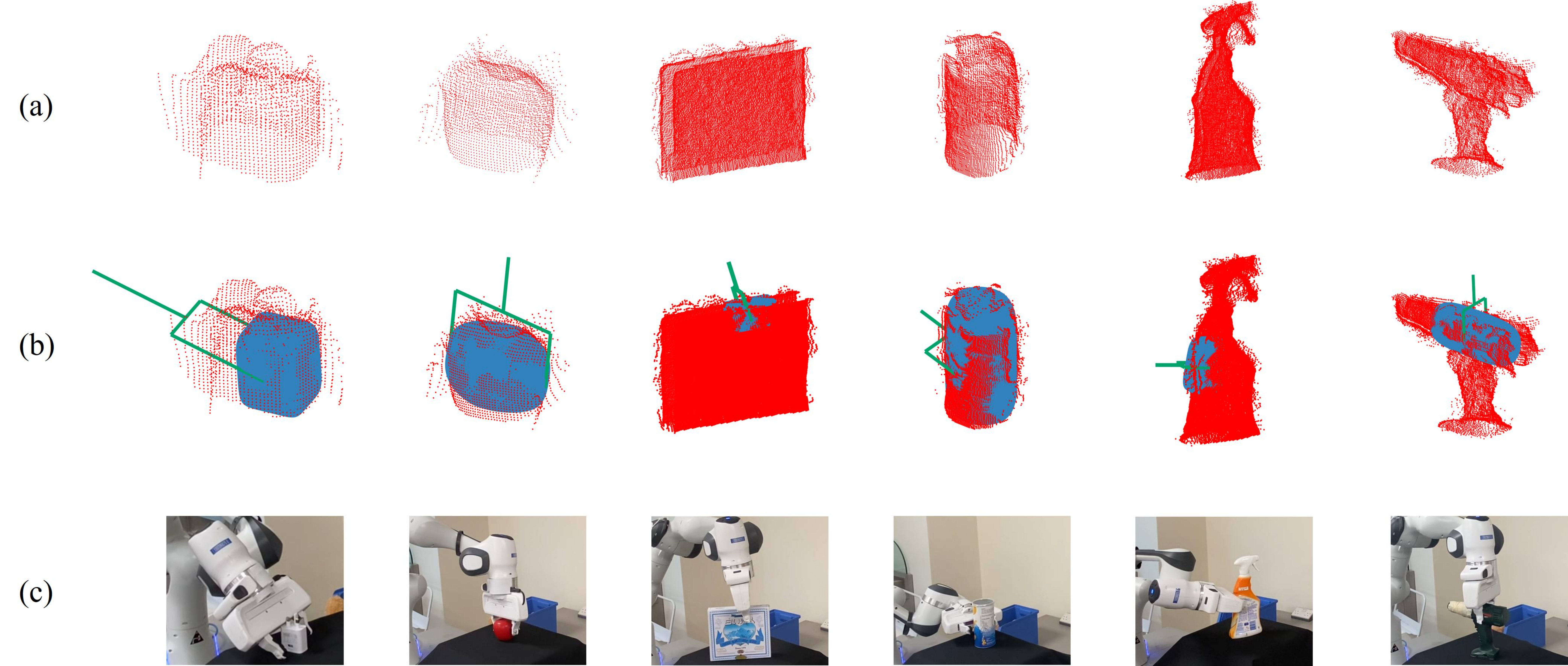}
    \caption{(a) Acquired point clouds of the objects. (b) grasping posed computed by our two-stage method. (3) Robot arm execution.}
    \label{fig:isolated_experiment}
\end{figure*}

We compute the GSR and compare it with results in other work, which is summarized in Table \ref{tab:isolated}.
Our method exhibits superior performance when grasping novel objects.
Our algorithm does not rely on any training datasets and predicts the grasping poses completely based on the geometric features of the objects.
Thus, our method can predict good grasping poses when the objects are placed at different poses.
Furthermore, noise resistance is noteworthy.
As shown in Fig \ref{fig:isolated_experiment}, the acquired point clouds from the sensors are noisy due to errors given rise to sensors and calibration, which are inevitable in real-life applications.
But, our method is capable of dealing with such errors.
The probabilistic model we adapt from \cite{Liu_2022_CVPR} can rule noisy points as outliers and ignore them when recovering superquadrics.
In addition, the optimization process can smooth out the errors brought by calibration.
Thus, our method is robust in the presence of noise and errors.
The final grasping pose can be calculated within 1 to 1.5 seconds for each object. 

\begin{table}[h]
    \setlength{\tabcolsep}{7pt} 
    \renewcommand{\arraystretch}{1.2}
    \centering
    \caption{Quantitative results of grasping previously unseen isolated household objects. We report Grasp Success Rate (GSR).}
    \begin{tabular}{cc}
    \hline
     Method & GSR (\%) \\
    \hline 
    GraspNet \cite{mousavian20196} & $88$ \\
    GDP \cite{ten2017grasp} & $47$  \\
    PointGDP \cite{liang2019pointnetgpd} & $82$ \\
    \hline
    ours & $96$ \\
    \hline
    \end{tabular}
    \label{tab:isolated}
\end{table}

\subsection{Packed Scenes}
\begin{figure*}
    \centering
    \includegraphics[width=1.9\columnwidth]{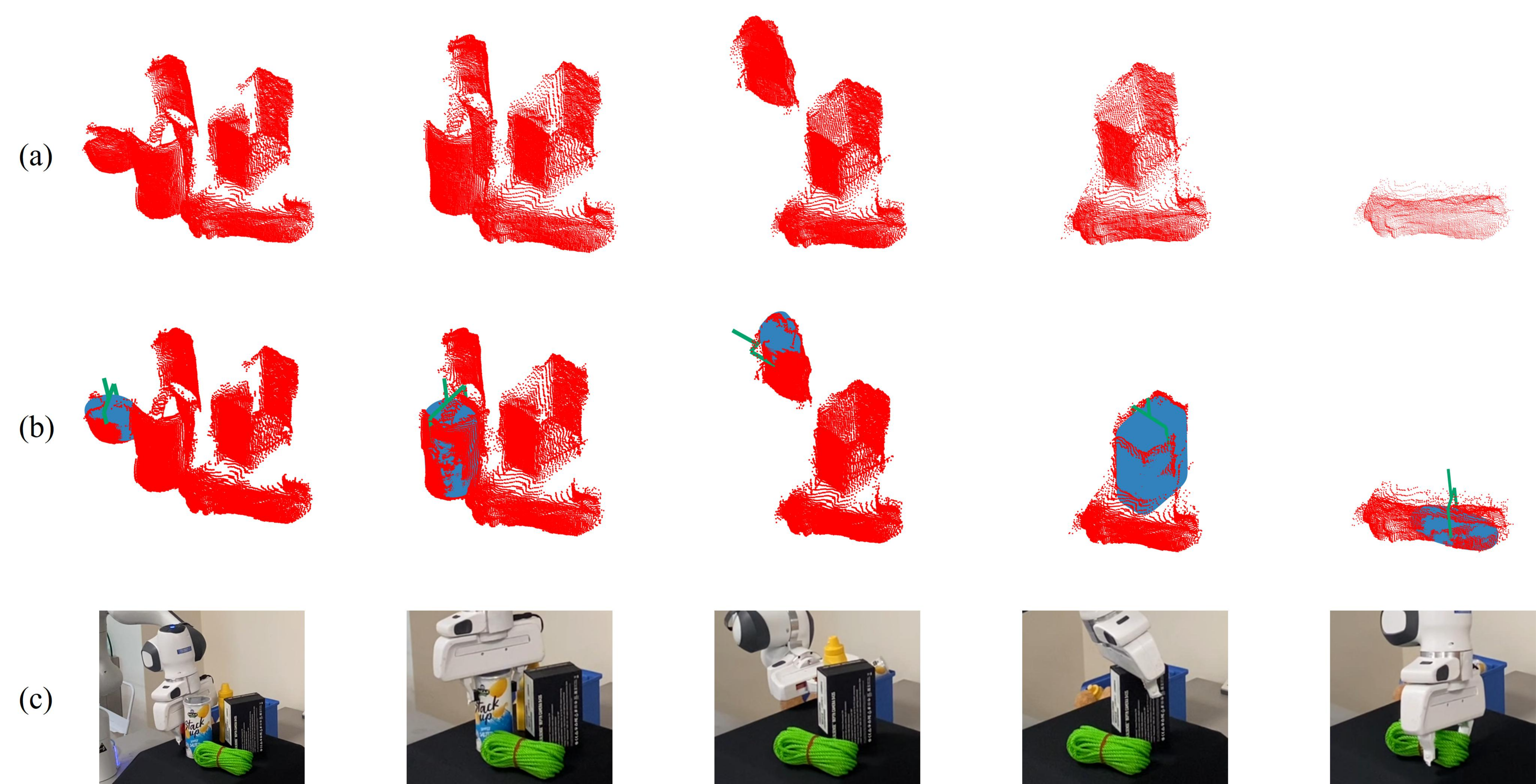}
    \caption{(a) Acquired point clouds of the packed scene. (b) grasping posed computed by our two-stage method. (3) Robot arm execution.}
    \label{fig:packed_experiment}
\end{figure*}

Even though our method is designed for isolated objects, we can easily extend it to packed scenes by recognizing the whole cluttered scene as a big single object.
After that, every step is exactly the same except we remove the score of distance to the center of mass since the center of mass of a cluttered scene will be meaningless.
At each round, we randomly choose 5 out of the 15 objects and place them at the scene.
Each round of decluttering terminates if there are no objects left in the scene or two consecutive attempts fail.
We conduct 15 rounds of experiments and report both GSR and CR and compare them with GIGA \cite{jiangsynergies}.
The results are summarized in Table \ref{tab:packed}.
Even in a cluttered scene, our algorithm can still identify the hidden superquadric among the noisy point clouds and predict the right grasping pose to remove objects one by one as shown in Fig. \ref{fig:packed_experiment}.

\begin{table}[h]
    \setlength{\tabcolsep}{7pt} 
    \renewcommand{\arraystretch}{1.2}
    \centering
    \caption{Quantitative results of packed scenes removal. We report Grasp Success Rate (GSR) and Clearance Rate (CR).}
    \begin{tabular}{lcc}
    \hline
     Method & \multicolumn{2}{c}{ Packed } \\
     & GSR (\%) & CR $(\%)$ \\
    \hline 
    GIGA\cite{jiangsynergies} & $83.3$ & $86.6$ \\
    ours & $81.3$ & $ 81.3$ \\ 
    \hline
    \end{tabular}
    \label{tab:packed}
\end{table}

\section{Conclusion}
To recapitulate, we propose a novel two-stage algorithm to synthesize grasping poses from the point clouds of a single object or a packed scene.
Our algorithm first identifies several hidden superquadrics inside the scene.
The adapted probabilistic model allows our method to resist noise resulting from depth sensors.
Subsequently, a list of grasping candidates is generated from the hidden superquadrics by exploiting the tri-symmetry of superquadrics.
After eliminating invalid poses from the candidate pools, a simple yet effective evaluation model quantifies the quality of each candidate and the one with the highest score will be selected as the final grasping pose.
As the experiments demonstrate, our method shows competitive performance on previously unseen objects and packed scenes compared with the state-of-the-art without big dataset and prior training.
As currently formulated, there are some limitations to our work. 
For starters, our approach needs to fix two sensors in space to obtain a relatively complete point cloud of the object to identify the hidden superquadrics. 
Secondly, it is difficult to isolate the object from the planar surface for thin objects like scissors or spoons.
And thus, our approach finds it hard to identify the hidden superquadrics correctly for those thin objects.
Thirdly, our method does not show better performance when it comes to packed scenes compared with the state-of-the-art.
We observe that many failure cases are caused by collisions with other objects when the robot arm attempts to move one of the objects from the scene to the bin.
Therefore, future work will focus on combining motion planning algorithms with our method to produce a collision-free grasp and a collision-free motion.
In addition, analyzing the functionality of the object together with geometry and then generating task-oriented grasps can be interesting in the future, as well.


\section*{Acknowledgments}
This work was supported by NUS Startup  grants A-0009059-02-00 and A-0009059-03-00, CDE Board Fund E-465-00-0009-01, National Research Foundation, Singapore, under its Medium Sized Centre Programme - Centre for Advanced Robotics Technology Innovation (CARTIN), 
subaward A-0009428-08-00, and AME Programmatic Fund Project MARIO A-0008449-01-00.


\bibliographystyle{plainnat}
\bibliography{references}

\end{document}